%% file: main.tex
\documentclass{article}

\PassOptionsToPackage{square, numbers, comma, sort&compress}{natbib}

\usepackage[preprint,nonatbib]{neurips_2025}

\usepackage[utf8]{inputenc} 
\usepackage[T1]{fontenc}    
\usepackage{hyperref}       
\usepackage{url}            
\usepackage{booktabs}       
\usepackage{amsfonts}       
\usepackage{xspace}         
\usepackage{nicefrac}       
\usepackage{microtype}
\usepackage{xcolor}
\usepackage{enumitem}

\usepackage{multirow}  
\usepackage{amsmath}
\usepackage{graphicx}
\usepackage{bm}
\usepackage{wrapfig}
\usepackage{marvosym}

\usepackage[capitalise]{cleveref}
\crefname{figure}{Fig.}{Figs.}
\crefname{table}{Tab.}{Tabs.}
\crefname{section}{Sec.}{Secs.}
\crefname{subsection}{Sec.}{Secs.}
\crefname{equation}{Eq.}{Eqs.}

\def\eg{\emph{e.g.}\xspace}

\newcommand{\method}{PM-Loss\xspace}

\newcommand{\boldstart}[1]{\vspace{0.1in}\noindent\textbf{#1}}

\title{Revisiting Depth Representations for Feed-Forward 3D Gaussian Splatting}

\author{
Duochao Shi\textsuperscript{$1*$}, \quad
Weijie Wang\textsuperscript{$1,4*$}, \quad
Donny Y. Chen\textsuperscript{$2$}, \quad
Zeyu Zhang\textsuperscript{$2,4$}, \\[.12cm]  
\textbf{Jia-Wang Bian\textsuperscript{$3$},} \quad
\textbf{Bohan Zhuang\textsuperscript{$1$},} \quad
\textbf{Chunhua Shen\textsuperscript{$1$}
} \\ 
{} 
\\
\textsuperscript{$1$}Zhejiang University, China
\quad
\textsuperscript{$2$}Monash University, Australia
\quad
\textsuperscript{$3$}MBZUAI
\quad
\textsuperscript{$4$}GigaAI
}

\begin{document}

\footnotetext[1]{\textsuperscript{$*$}Equal contribution.}

\maketitle

\input{arxiv_sections/0_abstract}

\input{arxiv_sections/1_introduction}

\input{arxiv_sections/2_related}
\input{arxiv_sections/3_method}

\input{arxiv_sections/4_experiments}

\input{arxiv_sections/5_conclusion}

\newpage

\bibliographystyle{unsrt}
\bibliography{main}

\newpage
\input{arxiv_sections/6_appendix}

\end{document}

%% file: arxiv_sections/0_abstract.tex
\begin{abstract}

Depth maps are widely used in feed-forward 3D Gaussian Splatting (3DGS) pipelines by unprojecting them into 3D point clouds for novel view synthesis. This approach offers advantages such as efficient training, the use of known camera poses, and accurate geometry estimation. However, depth discontinuities at object boundaries often lead to fragmented or sparse point clouds, degrading rendering quality---a well-known limitation of depth-based representations. To tackle this issue, we introduce \textbf{\method}, a novel regularization loss based on a pointmap predicted by a pre-trained transformer. Although  the pointmap itself may be less accurate than the depth map, it effectively enforces geometric smoothness, especially around object boundaries. With the improved depth map, our method significantly improves the feed-forward 3DGS across various architectures and scenes, delivering consistently better rendering results. Our project page: \url{https://aim-uofa.github.io/PMLoss}

\end{abstract}

%% file: arxiv_sections/1_introduction.tex
\section{Introduction}\label{sec:intro}

Novel view synthesis (NVS) is a long-standing topic in computer vision and graphics, recently drawing increasing attention due to advances in neural rendering, particularly 3D Gaussian Splatting (3DGS)~\cite{kerbl20233d}. While NVS models take 2D images as inputs and outputs, their primary goal is to recover the underlying 3D scene structure. Hence, smooth and accurate geometry is essential for generating high-quality novel views. This has led to a series of research efforts aimed at enhancing visual quality by learning more precise and consistent geometric representations~\cite{huang20242d,fan2024trim,lyu20243dgsr,wolf2024surface,yu2024gsdf}.

Although 3DGS models have ultra-fast rendering speed, reconstructing them for unseen scenes requires a time-consuming per-scene optimization process, limiting their usability in real-world applications. This challenge has led to the development of feed-forward 3DGS methods~\cite{charatan2024pixelsplat,chen2024mvsplat}, the main focus of our work. Unlike per-scene tuning methods that improve visual quality by learning better geometry, feed-forward 3DGS models typically fall short in geometric quality, despite significant progress aimed at enhancing  appearance~\cite{xu2024depthsplat,szymanowicz2024flash3d,ziwen2024long,zhang2024gs,wang2024freesplat,zhang2025transplat}. The core issue lies in the representation used by feed-forward methods, which rely on \emph{depth maps}. Most feed-forward models predict depth maps and then unproject them to form 3D Gaussians. Since depth maps often contain discontinuities near object boundaries~\cite{ramamonjisoa2020predicting,sun2023sc}, directly unprojecting them transfers these artifacts to the 3D representation, resulting in degraded geometry quality.

Recently, 3D reconstruction has been dominated by a new line of research that adopts a representation known as the pointmap~\cite{wang2024dust3r}. Unlike depth maps, which represent a scalar value $d \in \mathbb{R}^1$ in camera space, pointmaps encode a set of 3D points $p \in \mathbb{R}^3$  in world space, allowing for smoother and more accurate modeling of geometry. In addition, pointmaps simplify the traditional multi-view stereo (MVS)~\cite{yao2018mvsnet,gu2020cascade} process by reformulating it as a direct regression task through a neural network. These advantageous properties have contributed to the success of many recent feed-forward approaches to 3D reconstruction~\cite{yang2025fast3r,tang2024mv,zhang2024monst3r,wang2025vggt,wang2024moge,wang20243d}.

The success of pointmaps in accurate and regression-based 3D reconstruction motivates us to introduce them as a strong prior to reduce artifacts in depth map based feed-forward 3DGS. This is not straightforward, as pointmaps \emph{implicitly} encode \emph{coarse} camera poses~\cite{wang2024dust3r}, while feed-forward 3DGS performs best with \emph{explicitly} provided \emph{accurate} poses~\cite{charatan2024pixelsplat,chen2024mvsplat,ziwen2024long}, making it challenging to leverage the geometry prior effectively. Existing methods that adopt pointmap priors in feed-forward 3DGS often assume a pose-free setting~\cite{smart2024splatt3r,ye2024no}. While this avoids the pose issue by ignoring it, novel view evaluation either relies on testing with a specific dataset already used by the pointmap model (\eg, ScanNet~\cite{yeshwanth2023scannet++} in Splatt3R~\cite{smart2024splatt3r}) or requires slow test time pose alignment (\eg, NoPoSplat~\cite{ye2024no}), both of which hinder real world usability. Although one might inject camera poses into pointmap-based feed-forward models using, for example, Pl\"ucker ray embeddings, this approach is suboptimal as it requires expensive retraining to realign the pose distribution implicitly embedded in pointmaps and does not enhance the quality of scene details.

\begin{figure}
    \centering
    \vspace{0.3cm}
    \includegraphics[width=\linewidth]{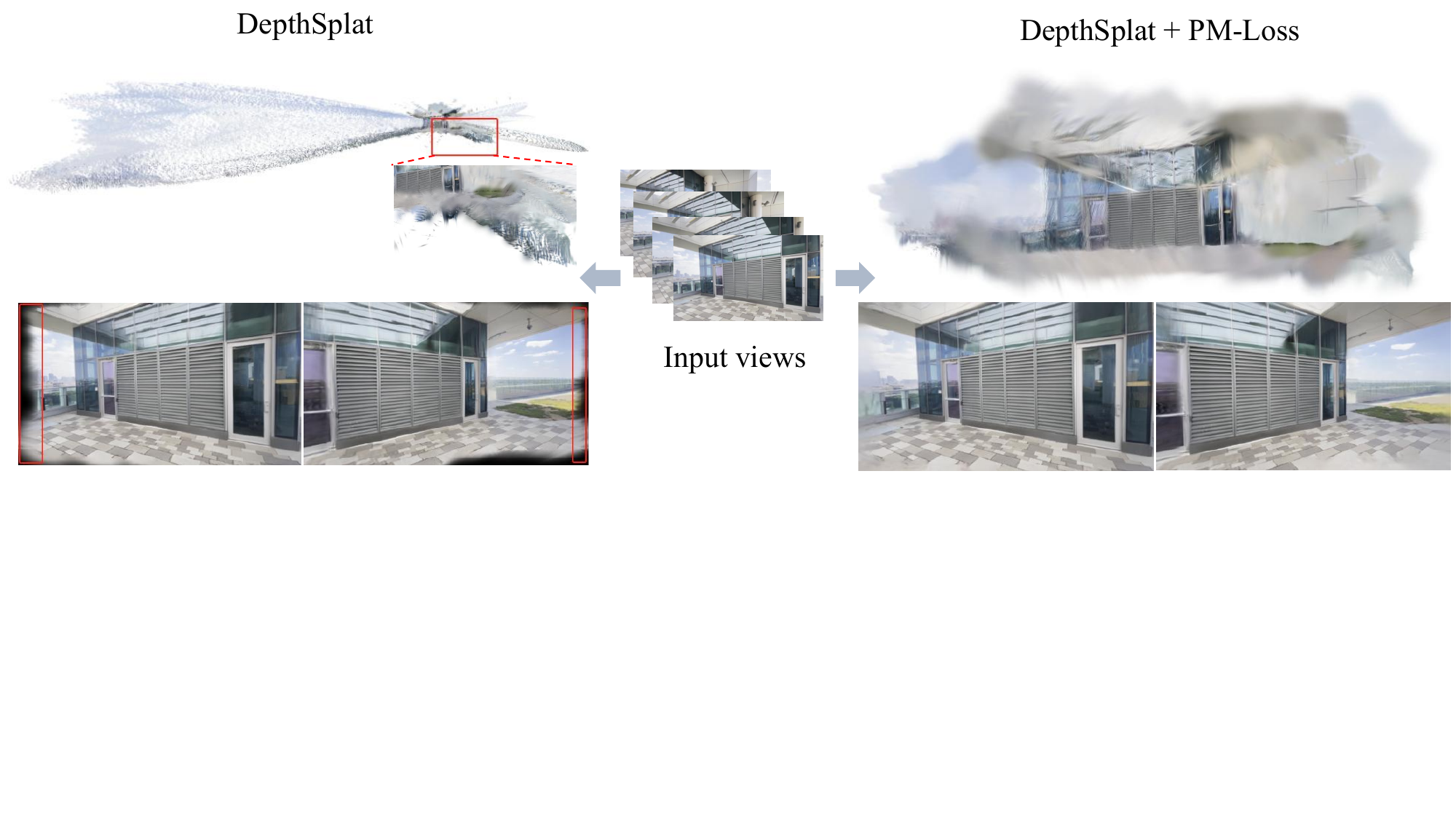}
    \vspace{-3.6cm}
    \caption{Feed-forward 3DGS models, \eg, DepthSplat~\cite{xu2024depthsplat}, rely on unprojected depth to form 3D Gaussians. The inherent discontinuities of depth near object boundaries often propagate into distorted 3D point clouds (top left) and degraded rendering (bottom left). Our \method addresses this by using the prior from pointmap, achieving higher-quality geometry (top right) and rendering (bottom right).}
    \label{fig:teaser}
    \vspace{-0.1cm}
\end{figure}

We propose a novel method to transfer the geometry prior from pointmap regression models to feed-forward 3DGS by formulating it as a simple yet effective training loss. Unlike prior methods~\cite{smart2024splatt3r,ye2024no} that attach an additional ``Gaussian head'' to the pointmap backbone, introducing the pose dilemma and requiring customization for each specific model, our approach is plug-and-play and avoids pose issues entirely.
In particular, our \method guides the learning of point clouds unprojected from predicted depth by taking the global pointmap predicted by a large-scale 3D reconstruction model, \eg, Fast3R~\cite{yang2025fast3r}, VGGT~\cite{wang2025vggt}, as a pseudo-ground truth. This guidance requires that the source and target points are in the same space and that efficient measurements are available. For the former, we find that the Umeyama algorithm can efficiently align the two point clouds (see \cref{tab:component_time}), leveraging the one-to-one correspondence between depth maps and pointmaps. For the latter, the Chamfer loss is used to directly regularize them in 3D space, leading to significantly better geometry quality than those applied in 2D space (see \cref{tab:pointmap_compare_pc}). By distilling the geometry prior embedded in the pointmap predicted by a pre-trained 3D reconstruction model, our method helps ease discontinuities caused by unprojected depth and significantly boosts the quality of predicted 3D point clouds and rendered novel views for feed-forward 3DGS models---See \cref{fig:teaser}.

To verify the effectiveness of our \method, we apply it to train two representative feed-forward 3DGS models, MVSplat~\cite{chen2024mvsplat} and DepthSplat~\cite{xu2024depthsplat}, on two large-scale benchmarks, RealEstate10K~\cite{zhou2018stereo} and DL3DV-10K~\cite{ling2024dl3dv}. Experiments demonstrate that our \method improves both the quality of the 3D Gaussians and the rendered novel views across all reported metrics. Extensive ablation studies and analysis further validate architectural design choices, as well as efficient memory and runtime usage of our \method. Given its plug-and-play, efficient, and effective nature, we believe that \method will play an important role in training feed-forward 3DGS in the future. Our contributions are threefold,

\begin{itemize}[leftmargin=*]
    \item We pinpoint an unexposed yet critical issue that leads to lower-quality 3D Gaussians predicted by feed-forward 3DGS models, rooted in the long-standing discontinuity issue of depth.
    \item We introduce a novel training loss, \method, designed to improve 3D Gaussian quality by leveraging the geometry prior from pointmaps obtained from pre-trained 3D reconstruction models.
    \item Extensive experiments on existing feed-forward 3DGS models across two large-scale datasets demonstrate the effectiveness of our \method in enhancing the quality of both 3D Gaussians and rendered novel views.    
\end{itemize}

%% file: arxiv_sections/2_related.tex
\section{Related Work}\label{sec:related}

\subsection{3D Gaussian Splatting}

Novel view synthesis (NVS) techniques have evolved significantly, transitioning from traditional image-based methods~\cite{seitz1996view, chen2023view} to modern neural rendering approaches~\cite{kerbl20233d, mildenhall2020nerfrepresentingscenesneural}. Early neural methods, such as NeRF~\cite{mildenhall2020nerfrepresentingscenesneural}, represent scenes implicitly. While they can achieve high-fidelity results, they are typically hampered by slow rendering speeds and the requirement for dense input views. More recently, explicit representations like 3D Gaussian Splatting (3DGS)~\cite{kerbl20233d} and its subsequent variants\cite{huang20242d,fan2024trim,lyu20243dgsr,wolf2024surface,yu2024gsdf} have emerged, offering significantly faster rendering speeds due to their rasterization-friendly nature.

Several recent works have explored leveraging geometric priors to optimize the 3D Gaussian representation. Specifically, some methods~\cite{chung2024depthregularizedoptimization3dgaussian, zhu2023FSGS} incorporate depth information as a prior for optimizing the geometry of the Gaussians. However, a common limitation of these approaches is their reliance on monocular depth estimation models~\cite{depth_anything_v1, depth_anything_v2}. Such monocular priors, especially when estimated independently per-image, often suffer from inconsistencies and multi-view misalignment, hindering geometric accuracy. Our approach mitigates this multi-view inconsistency by leveraging supervision from a multi-view consistent 3D point prior derived from pretrained models.

\subsection{Feed-forward 3D Gaussian Splatting}

Moving beyond per-scene optimization, pixelSplat~\cite{charatan2024pixelsplat} presented a pioneering feed-forward approach for 3DGS, predicting Gaussian parameters directly from two input views with help of epipolar transformers. Subsequently, MVSplat~\cite{chen2024mvsplat} improved the efficiency of feed-forward 3DGS by proposing a cost-volume based feature fusion method, enhancing its practicality. DepthSplat~\cite{xu2024depthsplat} further explored incorporating depth priors~\cite{depth_anything_v1, depth_anything_v2} into the feed-forward 3DGS framework, aiming to improve the geometric accuracy of the predictions. Same as the previous methods~\cite{wang2024freesplat, chung2024depthregularizedoptimization3dgaussian, zhu2023FSGS, wang2025freesplat++, min2024epipolarfree, fei2024pixelgaussian, kang2025selfsplat, wang2025zpressor, chen2024mvsplat360}, these feed-forward methods are often challenged by the difficulty in ensuring accurate multi-view consistency during feature processing or prior fusion, leading to geometric inaccuracies in the generated 3D Gaussian representation.

\subsection{3D Reconstruction using Pointmaps}

Recent advancements in 3D pointmap reconstruction methods~\cite{wang2024dust3r, tang2024mv, wang2025vggt, wang20243d, mast3r_arxiv24, zhang2025flare, cut3r} that produce highly accurate 3D point clouds have gained significant attention. Representative works in this area include DUSt3R~\cite{wang2024dust3r}, which utilizes a large Transformer-based model for robust multi-view feature fusion to generate dense 3D points. 
Building upon DUSt3R~\cite{wang2024dust3r}, MV-DUST3R~\cite{tang2024mv} further extends its capability to handle an arbitrary number of input views by facilitating information exchange across them, typically considering one as a reference view.
Fast3R~\cite{yang2025fast3r} focuses on highly efficient reconstruction, demonstrating the ability to process over 1000 images in a single forward pass. VGGT~\cite{wang2025vggt} infers key 3D attributes of a scene by combining features from models like DINO and DPT. While these methods excel in geometric reconstruction accuracy, they are typically not designed for direct novel view synthesis and often incur significant training costs. Our method, \method, aims to bridge the gap by combining the strengths of efficient feed-forward 3DGS networks and the accurate geometric priors provided by point-map-based large models, resulting in improved geometric quality for feed-forward NVS.

%% file: arxiv_sections/3_method.tex
\begin{figure}[ht]
    \centering
    \label{fig:zju-logo}
    \includegraphics[width=\linewidth]{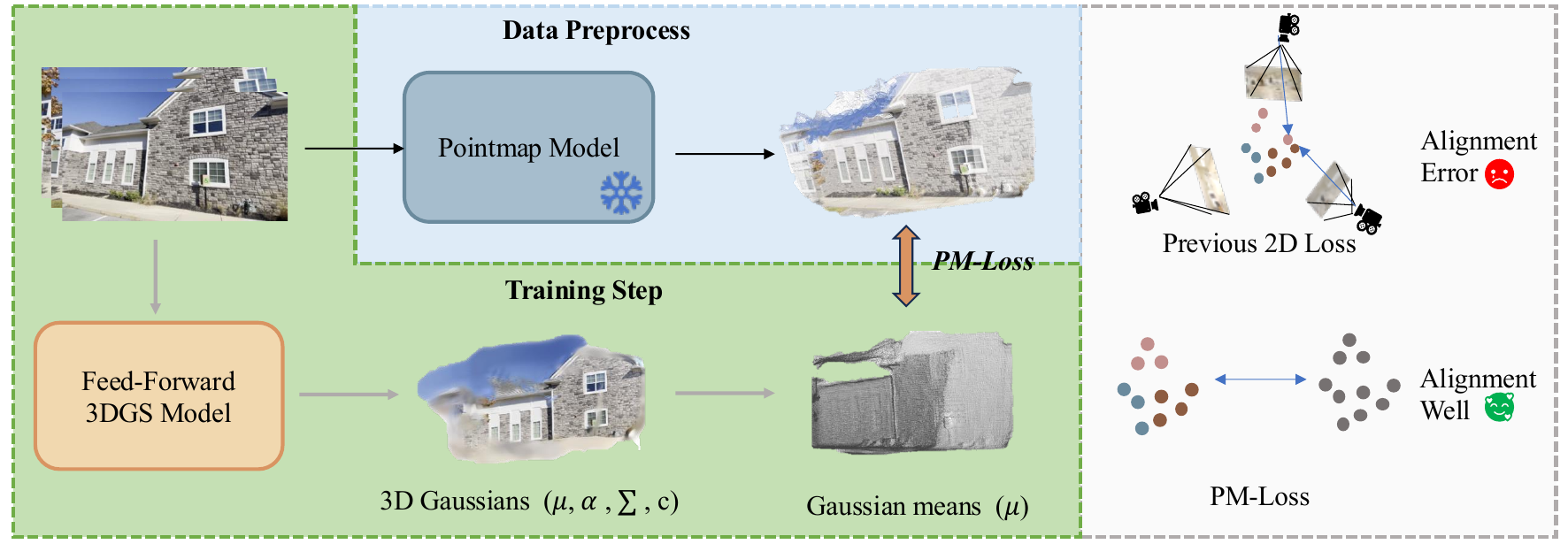}
    \vspace{-0.4cm}
    \caption{\textbf{Overview of \method.} The process begins by estimating a dense point map of the scene using a pre-trained model. This estimated point map then serves as direct 3D supervision for training a feed-forward 3D Gaussian Splatting model. Crucially, unlike conventional methods relying predominantly on 2D supervision, our approach leverages explicit 3D geometric cues, leading to enhanced 3D shape fidelity.}

\end{figure}

\section{Methodology}\label{sec:method}

Our goal is to train a neural network that directly predicts a 3D Gaussian Splatting (3DGS) model from one or more input images for novel view synthesis, eliminating the need for per-scene optimization.
To enhance the quality of the predicted 3D Gaussians, we introduce a novel PointMap Loss (\method) that regularizes the predicted 3D structure. \method leverages pointmaps—structured 2D-to-3D representations regressed from input images using a pretrained Vision Transformer\cite{vit}—to provide image-aligned supervision for geometry learning.
We begin by introducing the necessary preliminaries, followed by a detailed description of our \method design.

\subsection{Preliminary}
\label{subsec:preliminary}
\textbf{Feed-forward 3D Gaussian Splatting.}
The method aims to reconstruct a set of 3D Gaussians directly from one or several input images in a single forward pass. The general architecture involves an encoder-decoder structure. First, an encoder network processes the input image(s) $I$ to extract high-level features $F$:
\vspace{-0.1cm}
\begin{equation}
F = \text{Encoder}(I) \label{eq:encoder}
\end{equation}
These features $F$ are then typically fused with camera pose information $P_{\text{cam}}$ and potentially other supplementary information $S_{\text{aux}}$ through a fusion module, $\text{Fuse}(\cdot)$. A subsequent Gaussian head network, $G_{\text{head}}(\cdot)$, then predicts the parameters for a collection of $N$ 3D Gaussians. These parameters include the mean (center) $\mu \in \mathbb{R}^3$, covariance $\Sigma \in \mathbb{R}^{3 \times 3}$ (often represented by scale and rotation), opacity $\alpha \in \mathbb{R}$, and color $c \in \mathbb{R}^3$ (or spherical harmonic coefficients) for each Gaussian:
\begin{equation}
(\{\mu_i, \Sigma_i, \alpha_i, c_i\}_{i=1}^N) = G_{\text{head}}(\text{Fuse}(F, P_{\text{cam}}, S_{\text{aux}})) \label{eq:ghead}
\end{equation}
In typical feed-forward 3DGS pipelines, the Gaussian means $\mu_i$ are derived by unprojecting predicted depth maps. Specifically, for each pixel $(u,v)$ in an input image, a depth value $d$ is predicted. This depth, along with the camera intrinsic matrix $K$ and camera-to-world transformation matrix $T_{cw} = [R_{\text{ext}} | t_{\text{ext}}]$, is used to compute the 3D position of the corresponding Gaussian center $\mu_{uv}$:
\begin{equation}
\mathbf{p}_{\text{cam}}(u,v) = d(u,v) \cdot K^{-1} \begin{pmatrix} u \\ v \\ 1 \end{pmatrix} \label{eq:pixel_to_cam}
\end{equation}
\vspace{-0.1cm}
\begin{equation}
\mu_{uv} = R_{\text{ext}} \cdot \mathbf{p}_{\text{cam}}(u,v) + t_{\text{ext}} \label{eq:cam_to_world}
\end{equation}

While this approach is efficient, it often suffers from geometric inaccuracies. Depth maps inherently struggle with discontinuities, particularly around object boundaries. These discontinuities, when unprojected, lead to fragmented or misplaced Gaussians, thereby degrading the geometric quality of the 3D scene representation and subsequently impacting the novel view synthesis quality.

\textbf{Pointmap Regression.}
A pointmap is a structured 3D representation in which each pixel $(u,v)$ of an input 2D image $I$ is associated with a 3D point $p'_{uv} \in \mathbb{R}^3$ in world coordinates. Unlike depth maps, which provide only per-pixel Z-values, pointmaps directly represent full 3D coordinates (XYZ). They are typically regressed from images in a feed-forward manner using pretrained deep neural networks, often based on Vision Transformer (ViT)~\cite{vit} architectures.

Let $\mathcal{R}_{\text{pm}}$ denote such a pointmap regression model. For each of the $n_{\text{img}}$ input images $I_j$ (with resolution $H \times W$) and its camera pose $P_{\text{cam},j}$, $\mathcal{R}_{\text{pm}}$ outputs a set of 3D points:
\begin{equation}
\{p'_{j,u,v} \in \mathbb{R}^3 \mid (u,v) \text{ are pixel coordinates in } I_j\} = \mathcal{R}_{\text{pm}}(I_j, P_{\text{cam},j}) \label{eq:pointmap_regression}
\end{equation}
These per-image pointmaps are aggregated to form the global reference point cloud $X_{\text{PM}} = \{p'_k \in \mathbb{R}^3\}_{k=1}^{N_{\text{total\_pts}}}$, where $N_{\text{total\_pts}} = n_{\text{img}} \times H \times W$. This provides a dense 3D geometric prior that is leveraged in our \method.

\subsection{\method}
\label{subsec:pm_loss}

To address geometric inaccuracies in feed-forward 3DGS, existing methods such as DepthSplat~\cite{xu2024depthsplat} incorporate monocular depth priors. However, these priors are typically derived and supervised in 2D image space, which may not translate effectively into consistent 3D geometry.
Instead, we advocate for directly regularizing geometry learning in 3D space.

Given a batch of $n_{\text{img}}$ input images, each of resolution $H \times W$, the feed-forward 3DGS model aims to directly predict a set of 3D Gaussian centers.
We denote the collection of these predicted centers as $X_{\text{3DGS}}$, containing a total of $N_{\text{total\_pts}} = n_{\text{img}} \times H \times W$ points, where each point $\mu_k \in \mathbb{R}^3$  represents the center of a 3D Gaussian in world coordinates.

To guide the learning of accurate and consistent geometry, we introduce a 3D supervision signal derived from a pretrained pointmap regression model.
This model predicts a 3D point $p'_k \in \mathbb{R}^3$ for each pixel, resulting in a reference point cloud
$X_{\text{PM}}$ of the same cardinality.
Formally, we define the predicted and reference point sets as:
\begin{equation}
X_{\text{3DGS}} = \{\mu_k \in \mathbb{R}^3\}_{k=1}^{N_{\text{total\_pts}}}, \quad X_{\text{PM}} = \{p'_k \in \mathbb{R}^3\}_{k=1}^{N_{\text{total\_pts}}}
\label{eq:point_clouds}
\end{equation}
where $N_{\text{total\_pts}} = n_{\text{img}} \times H \times W$. 
Two point clouds, $X_{\text{3DGS}}$ and $X_{\text{PM}}$,
exhibit a natural one-to-one correspondence, as each $\mu_k$ and $p'_k$ originate from the same pixel $(i,j)$ of a specific input image $v$ within the batch. While the absolute accuracy of $X_{\text{PM}}$ might be less than that achievable via finely-tuned depth map unprojection in well-textured regions, the pointmaps tend to exhibit better geometric smoothness and completeness, especially near object boundaries.
We leverage this smoothness prior as a form of 3D supervision to guide the feed-forward 3DGS model toward learning more coherent and consistent geometry.

\textbf{Efficient point cloud alignment.} Although both $X_{\text{3DGS}}$ and $X_{\text{PM}}$ represent the scene’s 3D structure in world coordinates, directly supervising $X_{\text{3DGS}}$ using $X_{\text{PM}}$ is non-trivial. In practice, the two point clouds are often misaligned due to differences in scale, rotation, or translation—stemming from pose inaccuracies or implicit coordinate systems used by the pretrained model generating $X_{\text{PM}}$. Without addressing these discrepancies, point-wise supervision can introduce misleading gradients. Therefore, accurate alignment is crucial for effectively distilling the geometric prior from $X_{\text{PM}}$ into $X_{\text{3DGS}}$.

Traditional alignment methods such as Iterative Closest Point (ICP)\cite{besl1992method} are computationally expensive, particularly for dense point clouds, making them impractical for integration into the training loop.
However, in our case, both the Gaussian centers $X_{\text{3DGS}}$ (from per-pixel depth predictions) and the pointmap outputs $X_{\text{PM}}$ exhibit a one-to-one correspondence with input image pixels. Specifically, for each pixel $(u, v)$, there exists a matched pair $\mu_{uv} \in X_{\text{3DGS}}$ and $p'_{uv} \in X_{\text{PM}}$. Such natural correspondences allow us to apply the Umeyama algorithm~\cite{88573}, a closed-form and efficient solution for estimating the optimal similarity transformation (scale, rotation, and translation) between the two point sets.

Given $N_{\text{total\_pts}}$ corresponding points $X_{\text{PM}} = \{p'_k\}_{k=1}^{N_{\text{total\_pts}}}$ (source) and $X_{\text{3DGS}} = \{\mu_k\}_{k=1}^{N_{\text{total\_pts}}}$ (target), the Umeyama algorithm estimates the optimal scale factor $s^* \in \mathbb{R}^+$, rotation matrix $R^* \in SO(3)$, and translation vector $t^* \in \mathbb{R}^3$ by solving the following minimization problem:
\begin{equation}
(s^*, R^*, t^*) = \underset{s, R, t}{\text{argmin}} \frac{1}{N_{\text{total\_pts}}} \sum_{k=1}^{N_{\text{total\_pts}}} \|s R p'_k + t - \mu_k\|^2 \label{eq:umeyama_objective}
\end{equation}
The estimated transformation $(s^*, R^*, t^*)$ is then applied to each point $p'_k$ in the original pointmap $X_{\text{PM}}$ to obtain the aligned pointmap $X'_{\text{PM}} = \{s^* R^* p'_k + t^*\}_{k=1}^{N_{\text{total\_pts}}}$. 
This alignment enables us to compute the proposed supervision loss in a consistent coordinate frame.

\textbf{Single-directional Chamfer Loss.}
Given the aligned point clouds $X_{\text{3DGS}}$ and $X'_{\text{PM}}$,
we define the \method $L_{\text{PM}}$ as a single-directional Chamfer distance from $X_{\text{3DGS}}$ to $X'_{\text{PM}}$.
This formulation ensures that, for each point in $X_{\text{3DGS}}$, we can efficiently identify its nearest neighbor in $X'_{\text{PM}}$ to provide reliable geometric supervision.
Formally, the loss is defined as:
\begin{equation}
L_{\text{PM}}(X_{\text{3DGS}}, X'_{\text{PM}}) = \frac{1}{N_{\text{total\_pts}}} \sum_{\mu \in X_{\text{3DGS}}} \min_{p' \in X'_{\text{PM}}} \|\mu - p'\|_2^2 \label{eq:chamfer_loss}
\end{equation}
This formulation effectively acts as a regularisation term, penalizing deviations of the predicted Gaussian centers from the geometry prior suggested by the pointmap.
We adopt a simple mean squared error (MSE) averaged over all 3D Gaussian centers, which promotes stable training and ensures smooth gradient propagation.

A key insight of our proposed \method is to re-compute the nearest neighbors in 3D space for supervision, rather than directly relying on the natural one-to-one pixel correspondence, which degenerates to a depth loss. This design makes the supervision more robust to pose misalignments and prediction noise. 
We conduct ablation studies and report the quantitative results in Table~\ref{tab:pointmap_compare_pc}.

%% file: arxiv_sections/4_experiments.tex
\begin{figure}[t]
    \centering
    \includegraphics[width=\linewidth]{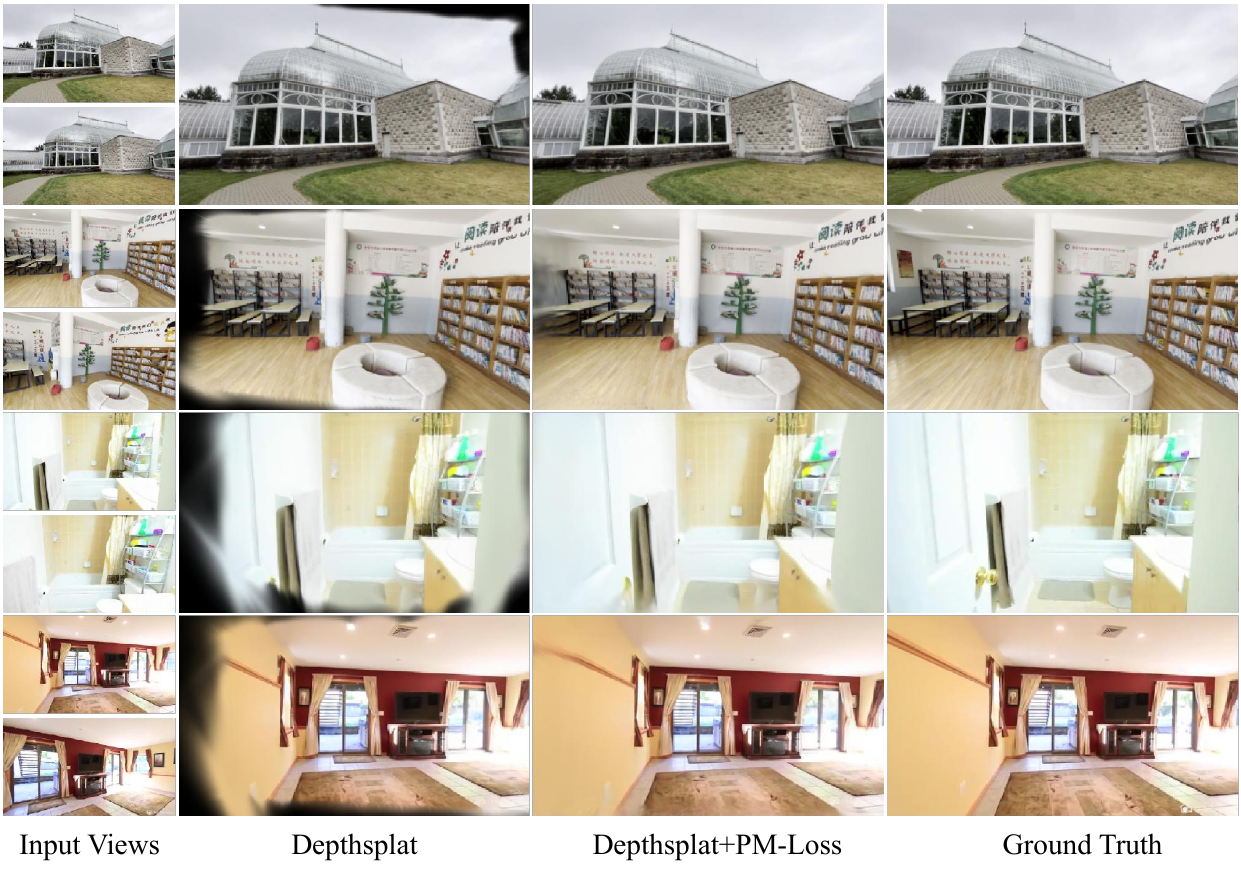}
    \vspace{-10.5cm}
    \caption{\textbf{Qualitative comparisons on DL3DV(top two rows) and RealEstate10K(bottom two rows) under the 2-view extrapolation setting. }Adding \method leads to significant improvements in rendering object boundaries.}
    \label{fig:extrapolation}
    \vspace{-0.2cm}
    
\end{figure}

\section{Experiments}
\label{sec:exp}

\subsection{Experimental Settings}
\label{subsec:settings}

\begin{table*}[t]
    \caption{\textbf{Quantitative comparisons on DL3DV under View Extrapolation. }Both MVSplat\cite{chen2024mvsplat} and DepthSplat\cite{xu2024depthsplat} show better rendering quality with the addition of \method. }
    \vspace{0.1cm}
    \centering
    \footnotesize
    \setlength{\tabcolsep}{2.5pt} 
    \begin{tabular}{@{}l ccc c ccc@{}}
    \toprule
    \multirow{2}{*}[-2pt]{\textbf{Method}} & \multicolumn{3}{c}{\textbf{DL3DV~\cite{ling2024dl3dv}}} && \multicolumn{3}{c}{\textbf{RealEstate10K~\cite{zhou2018stereo}}} \\
    \addlinespace[-12pt] \\
    \cmidrule{2-4} \cmidrule{6-8}
    \addlinespace[-12pt] \\
    & PSNR$\uparrow$ & SSIM$\uparrow$ & LPIPS$\downarrow$ && PSNR$\uparrow$ & SSIM$\uparrow$ & LPIPS$\downarrow$ \\
    \midrule

    DepthSplat & 18.46 & 0.689 & 0.261 && 20.43 & 0.788 & 0.218 \\
    DepthSplat+PM & \textbf{20.77} & \textbf{0.705} & \textbf{0.245} && \textbf{22.48} & \textbf{0.814} & \textbf{0.194} \\
    \midrule
    MVSplat & 16.79 & 0.592 & 0.322 && 19.52 & 0.757 & 0.231 \\
    MVSplat+PM & \textbf{19.25} & \textbf{0.615} & \textbf{0.291} && \textbf{22.18} & \textbf{0.787} & \textbf{0.199} \\
    \bottomrule
    \end{tabular}
    \label{tab:extrapolation}
\end{table*}

\begin{table}[t]
    \caption{\textbf{Quantitative comparison on DTU with varying input numbers. }Adding \method consistently improves geometry across different numbers of input views.}
    \centering
    \vspace{-0.1cm}
    \footnotesize
    \setlength{\tabcolsep}{4pt} 
    \begin{tabular}{@{}lccccccl@{}}
        \toprule
        \multirow{2}{*}{Input} & \multirow{2}{*}{Method} & \multicolumn{2}{c}{Acc} & \multicolumn{2}{c}{Comp} & \multicolumn{2}{c}{Overall} \\
        \cmidrule(lr){3-4} \cmidrule(lr){5-6} \cmidrule(lr){7-8}
        & & Mean & Med. & Mean & Med. & Mean & Med. \\
        \midrule
        \multirow{2}{*}{2-view} 
        & DepthSplat & 0.264 & 0.101 & 0.182 & 0.200 & 0.051 & 0.125 \\
        & DepthSplat+PM & \textbf{0.232} & \textbf{0.099} & \textbf{0.165} & \textbf{0.166} & \textbf{0.045} & \textbf{0.106} \\
        \midrule
        \multirow{2}{*}{4-view} 
        & DepthSplat & 0.169 & \textbf{0.066} & 0.123 & 0.117 & 0.022 & 0.051 \\
        & DepthSplat+PM & \textbf{0.156} & 0.069 & \textbf{0.113} & \textbf{0.076} & \textbf{0.022} & \textbf{0.049} \\
        \midrule
        \multirow{2}{*}{6-view} 
        & DepthSplat & 0.162 & \textbf{0.048} & 0.105 & 0.070 & 0.017 & 0.044 \\
        & DepthSplat+PM & \textbf{0.150} & 0.053 & \textbf{0.102} & \textbf{0.068} & \textbf{0.016} & \textbf{0.042} \\
        \bottomrule
    \end{tabular}
    \vspace{-0.2cm}
\label{tab:multi_view_dtu}
\end{table}

\textbf{Datasets.}
We evaluated our method using three datasets: DL3DV, RealEstate10K, and DTU.
DL3DV \cite{ling2024dl3dv} is a challenging large-scale collection comprising 10,510 real-world scenes. 
Following~\cite{xu2024depthsplat}, we used the DL3DV-Benchmark (140 scenes) for Novel View Synthesis (NVS) testing and the remaining DL3DV scenes for training.
RealEstate10K\cite{zhou2018stereo}, comprising camera trajectories from ~80,000 YouTube video clips (10M frames), is split into 67,477 training and 7,289 testing scenes. We used its test split for NVS evaluation, excluding scenes with too few views.
DTU \cite{jensen2014large} features 128 scenes from controlled lab environments with ground truth models from a structured light scanner and corresponding depth maps. We assessed Gaussian splat geometric quality on 16 of these scenes, following \cite{chen2023matchnerf} and \cite{yao2018mvsnet}.

\textbf{Baselines.}
To evaluate our proposed method, we apply it to two representative feed-forward 3D Gaussian Splatting (3DGS) models: MVSplat~\cite{chen2024mvsplat} and DepthSplat~\cite{xu2024depthsplat}. 
Our experiments compare the performance of models fine-tuned with \method against that of the same models fine-tuned with their original training objectives, ensuring both are evaluated in a fine-tuned state. For a fair comparison under otherwise identical conditions, both sets of models started from public pre-trained weights and were subsequently fine-tuned on the DL3DV dataset using the same batch size and total number of training iterations.

\textbf{Metrics.}
For Novel View Synthesis (NVS) evaluation, we focused on extrapolation scenarios to assess improvements in geometric quality, particularly at object edges. In these tests, two images served as context views. The target view was selected from outside the spatial region covered by these context views. This setup often results in target views near scene or object boundaries, which are challenging and help to demonstrate enhancements in edge geometry and rendering quality when applying our \method. Table~\ref{tab:extrapolation} presents the results of these extrapolation tests. 
In addition to NVS, to quantitatively assess the geometric quality of the generated 3D Gaussians, we treat the centers ($\mu$) of all 3D Gaussians as a point cloud and compare this representation against the ground truth (GT) point clouds provided by DTU. We use three standard point cloud metrics: Accuracy (Acc), Completeness (Comp), and the Overall Chamfer Distance (Overall), all where lower values indicate better performance. For each of these metrics, we report both the Mean and Median (Med.) values, see in ~\ref{tab:multi_view_dtu}.

\begin{figure}[t]
    \centering
    \includegraphics[width=\linewidth]{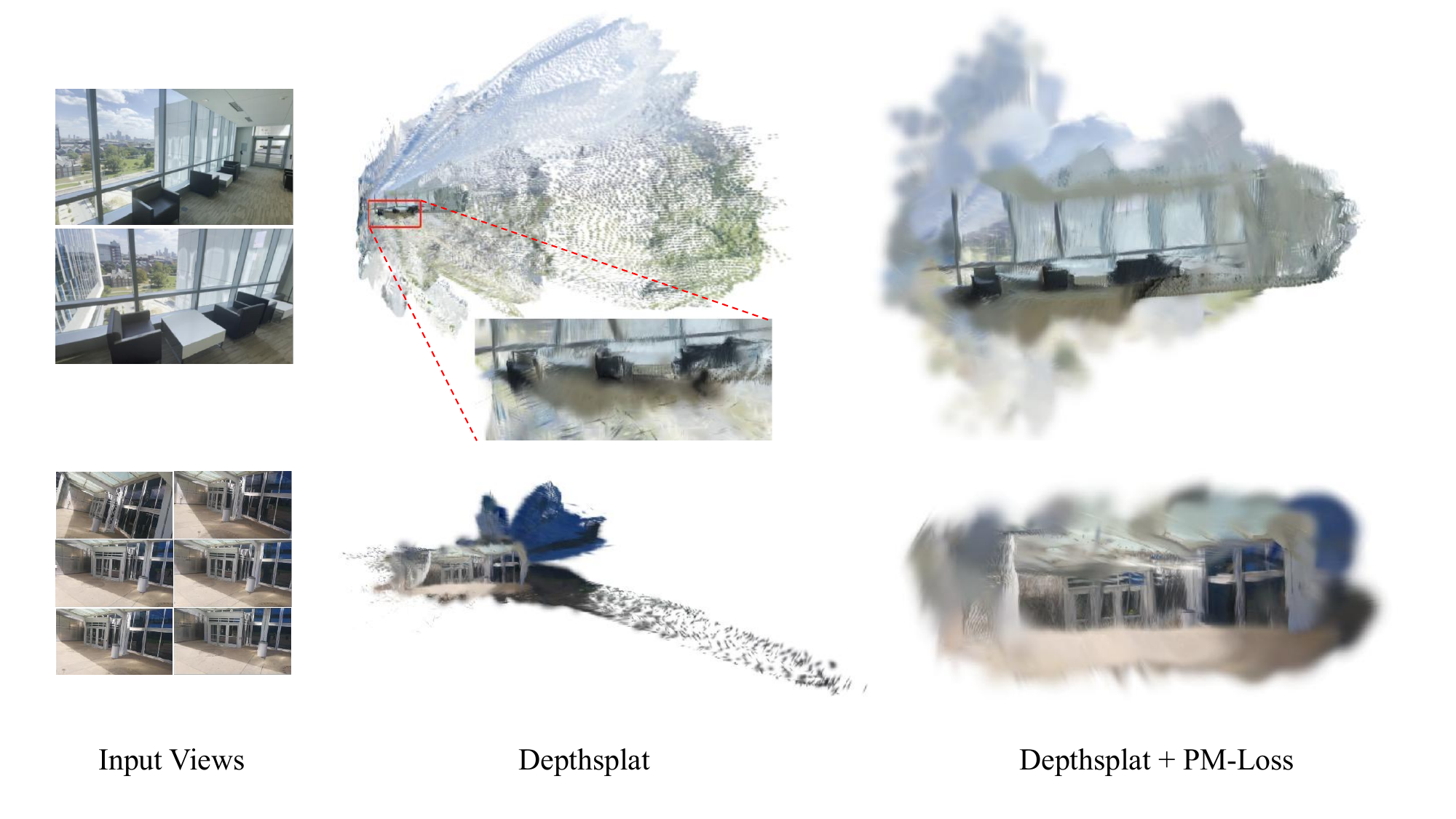}
    \vspace{-0.3cm}
    \caption{\textbf{Qualitative comparison of unprojected 3D Gaussians on DL3DV dataset. }Our method effectively constrains the 3D Gaussians, significantly reducing floating artifacts and noise near border.}
    \label{fig:gs_dl3dv}
\end{figure}

\begin{figure}[ht]
    \centering
    \includegraphics[width=\linewidth]{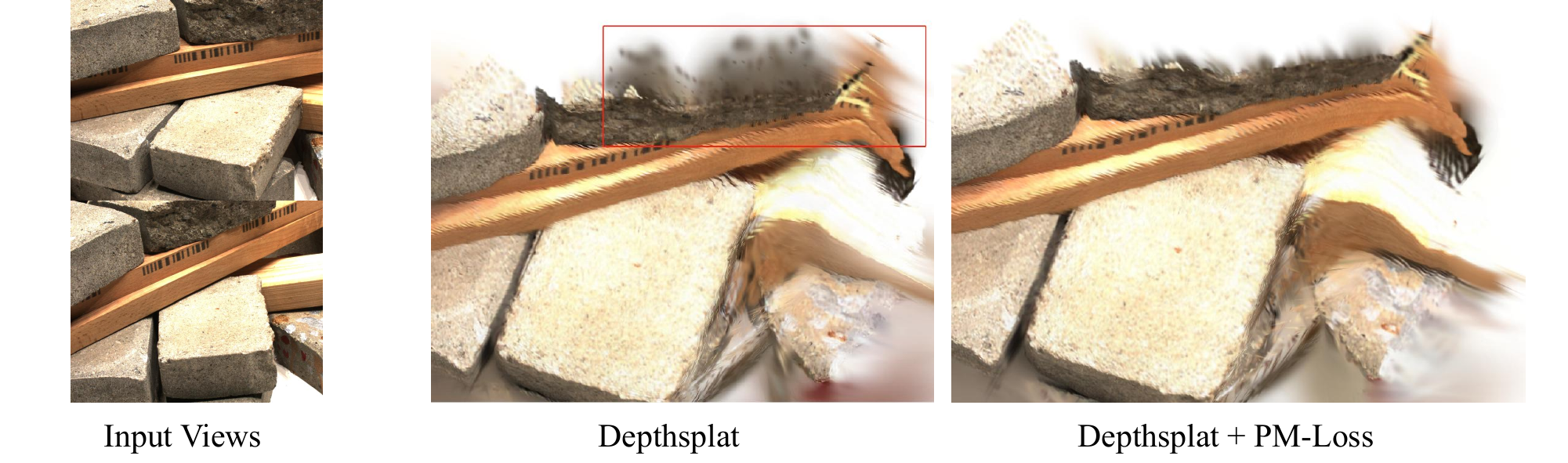}
    \vspace{-0.2cm}
    \caption{Qualitative comparison of unprojected 3D Gaussians on the DTU dataset.}
    \label{fig:gs_dtu}
    \vspace{-0.2cm}
\end{figure}

\textbf{Implementation Details.}
Our method was implemented in PyTorch. The Chamfer distance computation within our proposed \method utilizes the implementation from PyTorch3D~\cite{ravi2020pytorch3d}. All experiments were conducted on a single NVIDIA A100 GPU. We used the AdamW optimizer, fine-tuning each model variant (i.e., original base models and \method enhanced models) for $100,000$ iterations with a learning rate of $2 \times 10^{-4}$.
For the pointmap supervision component of \method, we adopted VGGT~\cite{wang2025vggt} to generate pseudo ground truth, specifically using its publicly available 1B parameter model. The weighting coefficient for \method, $\lambda_{PM}$, was set to $0.005$. Training and testing resolutions adhered to the original configurations of the base models: $256 \times 448$ for DepthSplat and $256 \times 256$ for MVSplat. Further training details are provided in the Appendix.

\subsection{Comparisons and Analysis}
\label{subsec:analysis}
In this section, we present detailed experimental results and analysis to verify the assumptions and effectiveness of our \method, covering rendered view quality, 3D point cloud quality, alternative design choices, memory and time efficiency of each component.

\textbf{Analysis of visual quality improvement.} 
By regularizing the predicted point clouds, our \method helps achieve better 3D Gaussian quality, which in turn leads to improved rendered novel view quality. As reported in \cref{tab:extrapolation}, it is clearly evident that our method boosts the performance of all baseline models on two large-scale datasets, with a consistent gain of at least 2 dB in PSNR. We hypothesize that this improvement comes mainly from the boundary regions of the scene, as our loss is particularly effective in addressing discontinuities around the boundaries. This is further supported by the qualitative results shown in \cref{fig:extrapolation}. Although baseline models render central regions of the scenes reasonably well, they often fail to reconstruct scene boundaries due to discontinuities rooted in the depth representation, resulting in black regions in the rendered views. In contrast, our \method provides additional regularization in 3D space by leveraging the geometry prior from the pointmap, which does not suffer from discontinuous boundaries. This enables a more accurate recovery of those regions and significantly improves the visual quality of rendered views from designated viewpoints.

\begin{table*}[t]
    \centering
    \caption{\textbf{Quantitative comparison of different distance measurements.} Our 3D ``nearest-neighbors'' Chamfer loss outperforms the 2D ``one-to-one'' depth loss, highlighting the benefits of 3D regularization.}
    \vspace{0.1cm}
    \small
    \begin{tabular}{@{}lccccccl@{}}
        \toprule
        \multirow{2}{*}[-2pt]{\textbf{Distance Measurement}} & \multicolumn{2}{c}{\textbf{Acc$\downarrow$}} & \multicolumn{2}{c}{\textbf{Comp$\downarrow$}} & \multicolumn{2}{c}{\textbf{Overall$\downarrow$}} \\
        \addlinespace[-12pt] \\
        \cmidrule(lr){2-3} \cmidrule(lr){4-5} \cmidrule(lr){6-7}
        \addlinespace[-12pt] \\
        & Mean & Med. & Mean & Med. & Mean & Med. \\
        \midrule
        2D ``one-to-one'' depth loss & 0.254 & \textbf{0.096} & 0.175 & 0.179 & 0.048 & 0.114 \\
        3D ``nearest-neighbors'' chamfer loss & \textbf{0.232} & 0.099 & \textbf{0.165} & \textbf{0.166} & \textbf{0.045} & \textbf{0.106} \\
        \bottomrule
    \end{tabular}
    \vspace{-3mm}
    \label{tab:pointmap_compare_pc}
\end{table*}

\begin{table}[h]
    \centering
    \begin{minipage}[t]{0.6\linewidth}
        \centering
        \footnotesize
        \caption{\textbf{Quantitative comparison of different pointmaps.} Higher quality pointmaps, such as VGGT, notably improve \method's effectiveness, though our approach performs well across different pointmap sources.}
        \label{tab:pointmap_selection}
        \setlength{\tabcolsep}{2.5pt} 
        \renewcommand{\arraystretch}{1.2}
        \begin{tabular}{@{}l ccc c ccc@{}}
        \toprule
        \multirow{2}{*}[-2pt]{\textbf{pointmap}} & \multicolumn{3}{c}{\textbf{DL3DV~\cite{ling2024dl3dv}}} && \multicolumn{3}{c}{\textbf{RealEstate10K~\cite{zhou2018stereo}}} \\
        \addlinespace[-12pt] \\
        \cmidrule{2-4} \cmidrule{6-8}
        \addlinespace[-12pt] \\
        & PSNR$\uparrow$ & SSIM$\uparrow$ & LPIPS$\downarrow$ && PSNR$\uparrow$ & SSIM$\uparrow$ & LPIPS$\downarrow$ \\
        \midrule
        w/o pointmap & 18.46 & 0.689 & 0.261 && 20.43 & 0.788 & 0.218 \\
        Fast3R~\cite{yang2025fast3r} & 20.51 & 0.690 & 0.257 && 22.43 & 0.812 & 0.197 \\
        VGGT~\cite{wang2025vggt} & \textbf{20.77} & \textbf{0.705} & \textbf{0.245} && \textbf{22.48} & \textbf{0.814} & \textbf{0.194} \\
        \bottomrule
        \end{tabular}
    \end{minipage}
    \hfill 
    \begin{minipage}[t]{0.3\linewidth}
        \centering
        \footnotesize
        \caption{\textbf{Timing breakdown of \method.} Our \method is efficient, with a total time of 65.0 ms for a typical operation.}
        \label{tab:component_time}
        \vspace{0.1cm} 
        \setlength{\tabcolsep}{4pt}
        \renewcommand{\arraystretch}{1.2}
        \begin{tabular}{@{}lc@{}}
            \toprule
            Component & Time (ms) \\
            \midrule
            Alignment (Umeyama) & 0.9 \\
            Chamfer Distance & 64.1 \\
            \midrule
            \textbf{Total} & \textbf{65.0} \\
            \bottomrule
        \end{tabular}
    \end{minipage}
\end{table}

\textbf{Analysis of point cloud quality improvement.}
We present a qualitative comparison of point cloud quality against DepthSplat on DL3DV in \cref{fig:gs_dl3dv}. The results support our motivation. DepthSplat produces 3D Gaussians with noisy floating-point artifacts around the border, while our \method yields sharper and cleaner borders. Since DL3DV contains only real-world scenes without ground truth point clouds, we also evaluate on a standard multi-view stereo benchmark that provides ground truth point clouds and depth maps. As reported in \cref{tab:multi_view_dtu} and shown in \cref{fig:gs_dtu}, the quality of the point cloud in terms of accuracy, completeness, and overall scores in DTU matches our findings on DL3DV and RE10K. Additionally, since DepthSplat can be evaluated with varying numbers of input views,  we perform similar tests using inputs ranging from 2 to 6 views. The consistent improvements brought by our \method further confirm its robustness and broad applicability.

\textbf{Impact of varying distance measurement.}
As stated in \cref{subsec:pm_loss}, one key insight of our \method is applying the regularization in 3D space, avoiding direct use of the one-to-one correspondence between pointmap and depth, which could potentially degenerate the regularization to a 2D depth loss. We demonstrate the effectiveness of this design choice in \cref{tab:pointmap_compare_pc}, compared to the 2D pointmap loss used in VGGT or Fast3R (excluding the confidence-aware component, as our backbone does not predict a confidence mask). Our approach of regularizing in 3D space (``nearest-neighbors'' Chamfer loss) consistently outperforms the 2D alternative (``one-to-one'' depth loss) across several challenging point cloud metrics. We hypothesize that this gain comes from the Chamfer loss computing distances based on nearest neighbors in world space, rather than relying on pixel-aligned correspondence in camera space. This makes it more robust to slight pose misalignments and leads to better effectiveness.

\textbf{Impact of varying pointmaps.}
Since our \method targets distilling the geometry prior knowledge from predicted point maps, one might wonder how the quality of pointmaps affects its effectiveness. We conduct experiments with alternative pointmap sources, comparing our default choice VGGT with another recent state-of-the-art counterpart, Fast3R. As reported in \cref{tab:pointmap_selection}, using pointmaps from Fast3R yields slightly lower performance on both datasets, while still significantly outperforming the baseline without pointmap regularization. This suggests that our \method is not tied to the VGGT architecture, although higher quality pointmaps do improve its effectiveness. We also observe a larger gain from VGGT over Fast3R on DL3DV (+0.2dB PSNR) than on RE10K (+0.05dB PSNR), indicating that our \method is particularly effective in more complex real-world scenarios.

\textbf{Efficiency of the default components.}
The extra computation cost of our \method mainly comes from two components, point cloud alignment and Chamfer loss calculation. Although we use pointmap as pseudo ground truth, it can be efficiently extracted offline (around 0.3s/scene for VGGT) and loaded through the dataloader as training input, so it does not count toward the total training cost. As reported in \cref{tab:component_time}, our \method introduces only a minor time overhead of about 60ms for a large volume of 458,752 3D Gaussians
(4 views with $256 \times 448$ resolution and pixel-aligned prediction)
, making it efficient to integrate into most existing feed-forward 3DGS models. This efficiency is partly due to the point-to-point correspondence between the pointmap and the depth-unprojected point clouds, which allows fast alignment using Umeyama\cite{88573}.

Without such an inherent correspondence, alignment would become significantly more challenging and costly, \eg, using ICP\cite{besl1992method} would take around 250ms for the same number of 3D Gaussians (see  \cref{tab:alignment_time}). This further supports the choice of using pointmap as our geometry prior.
Regarding memory, our \method increases training VRAM by 6.09GB \cref{tab:memory}. This additional VRAM is largely consumed by the VGGT-1B model during pointmap generation. Crucially, this generation is a preprocessing step. It can be performed entirely offline, meaning the VGGT-1B model itself would not need to occupy VRAM during the actual training process; only the resulting pointmap data would be loaded.
Furthermore, our \method is purely a regularization of training time and does \emph{not} introduce \emph{any} additional cost during testing.

\begin{table}[t]
    \centering
    \begin{minipage}[t]{0.45\linewidth}
        \centering
        \vspace{0.6cm}
        \footnotesize
        \caption{\textbf{Computation time comparison of alignment methods.} Our alignment method (Umeyama) is significantly more efficient than the commonly used ICP method.}
        \vspace{-2pt}
        \setlength{\tabcolsep}{4pt}
        \renewcommand{\arraystretch}{1.2}
        \begin{tabular}{@{}lc@{}}
            \toprule
            Method & Time (ms) \\
            \midrule
            Umeyama & 0.9 \\
            ICP & 238.3 \\
            \bottomrule
        \end{tabular}
        \label{tab:alignment_time}
    \end{minipage}
    \hfill 
    \begin{minipage}[t]{0.45\linewidth}
        \centering
        \footnotesize
        \vspace{0.6cm}
        \caption{\textbf{Memory usage of \method.} We report our max GPU memory usage during training DepthSplat with VGGT-1B pointmap model.}
        \setlength{\tabcolsep}{4pt}
        \renewcommand{\arraystretch}{1.2}
        \begin{tabular}{@{}lc@{}}
            \toprule
            Method & \begin{tabular}{@{}c@{}}Max VRAM (GB)\end{tabular} \\
            \midrule
            DepthSplat & 51.79 \\
            DepthSplat+PM & 57.88 \\
            \bottomrule
        \end{tabular}
        \label{tab:memory}
    \end{minipage}
    
\end{table}

%% file: arxiv_sections/5_conclusion.tex
\section{Conclusion}\label{sec:conclusion}
We present \method, a simple yet effective training loss that leverages geometry priors from pointmaps to improve feed-forward 3DGS. By regularizing in 3D space using global pointmaps as pseudo ground truth, \method alleviates depth-induced discontinuities near object boundaries, leading to significantly improved geometry and rendering quality.  
Our \method can be seamlessly integrated into existing training pipelines and introduce \emph{no} inference overhead. Extensive experiments and analysis on multiple backbones and large-scale datasets demonstrate its broad applicability and efficiency. We believe \method offers a practical solution for training more robust and accurate feed-forward 3DGS models.

\boldstart{Limitation and discussion.} The effectiveness of our \method is bounded by the quality of the pre-trained pointmap model, as errors in the pointmap may propagate into the feed-forward 3DGS model through our loss. Leveraging stronger pointmap models from future 3D reconstruction advances is a promising direction.

%% file: arxiv_sections/6_appendix.tex
\setcounter{table}{0}
\setcounter{figure}{0}
\setcounter{section}{0}
\renewcommand{\thetable}{\Alph{table}}
\renewcommand{\thefigure}{\Alph{figure}}
\renewcommand{\thesection}{\Alph{section}}

\section{More Experimental Analysis}
\label{sec:app_exp}

\textbf{Evaluation under view interpolation settings.}
\label{subsec:app_interpolation_eval}
Beyond the main paper's extrapolation experiments, we also tested our \method in view interpolation settings, where target views lie between the two context views. These tests used the DepthSplat's~\cite{xu2024depthsplat} offical evaluation set on DL3DV dataset, with two input views.
\cref{tab:interpolation} shows the quantitative results for this setup. Applying our \method (denoted as `+PM') consistently improves novel view synthesis for both DepthSplat~\cite{xu2024depthsplat} and MVSplat~\cite{chen2024mvsplat} across all metrics. This demonstrates that our distilled geometric priors are also beneficial for interpolated views.

Despite these positive interpolation results, our main paper primarily focuses on extrapolation. Extrapolation scenarios more rigorously test the geometric integrity of 3D reconstructions, especially near scene or object boundaries. This focus helps to clearly quantify and showcase improvements in 3D Gaussian geometric quality, a key contribution of our \method.

\begin{table*}[h]
    \caption{\textbf{NVS on DL3DV with DepthSplat~\cite{xu2024depthsplat}'s offical eval set. }Our \method (`+PM') improves rendering for both MVSplat~\cite{chen2024mvsplat} and DepthSplat~\cite{xu2024depthsplat}.}
    \vspace{0.1cm}
    \centering
    \footnotesize
    \setlength{\tabcolsep}{2.5pt} 
    \begin{tabular}{@{}l ccc@{}}
    \toprule
    \multirow{2}{*}[-2pt]{\textbf{Method}} & \multicolumn{3}{c}{\textbf{View Interpolation}} \\
    \addlinespace[-12pt] \\
    \cmidrule{2-4}
    \addlinespace[-12pt] \\
    & PSNR$\uparrow$ & SSIM$\uparrow$ & LPIPS$\downarrow$ \\
    \midrule
    DepthSplat & 19.05 & 0.610 & 0.313 \\
    DepthSplat+PM & \textbf{19.86} & \textbf{0.641} & \textbf{0.298} \\
    \midrule
    MVSplat & 17.59 & 0.514 & 0.396 \\
    MVSplat+PM & \textbf{17.81} & \textbf{0.520} & \textbf{0.380} \\
    \bottomrule
    \end{tabular}
    \label{tab:interpolation}
\end{table*}

\textbf{Memory efficiency with offline pointmap preprocessing}
\label{subsec:app_memory_offline}
As the main paper (Sec. 4.2) notes, pointmap generation is best performed as an offline preprocessing step. We analyze our \method's training VRAM footprint under this optimal workflow here. \Cref{tab:memory_offline} includes data for both offline and online pointmap processing scenarios when applying our \method (denoted as `+PM') to DepthSplat. 
Focusing on the recommended offline approach—where the pointmap generator's parameters do not consume training VRAM—our \method adds only a modest 0.96GB to DepthSplat's VRAM usage (52.75GB vs. 51.79GB). This small overhead, even when handling a large number of 3D Gaussians (e.g., approximately $4 \times 10^6$), primarily accounts for storing the loaded pointmap data and the computations for our online loss components (alignment and Chamfer distance). This highlights the efficiency of our core \method component.

\begin{table}[h]
    \centering
    \caption{\textbf{VRAM footprint of different pointmap processing strategies.} Comparison of maximum GPU memory for baseline DepthSplat, and DepthSplat with \method (offline vs. online VGGT-1B pointmap processing) during training.}
    \setlength{\tabcolsep}{4pt}
    \renewcommand{\arraystretch}{1.2}
    \begin{tabular}{@{}lc@{}}
    \toprule
    Method & \begin{tabular}{@{}c@{}}Max VRAM (GB)\end{tabular} \\
    \midrule
    DepthSplat & 51.79 \\
    DepthSplat+PM (offline process) & 52.75 \\
    DepthSplat+PM (online process) & 57.88 \\
    \bottomrule
    \end{tabular}
    \label{tab:memory_offline}
\end{table}

\textbf{Discussion of existing pointmap based feed-forward 3DGS.}
\label{subsec:app_discussion_integration_strategies}
Integrating pointmap priors into feed-forward 3DGS faces challenges from pose discrepancies (Discussed in Sec. 1). Some methods, like NoPoSplat~\cite{ye2024no}, use specialized ``Gaussian heads'' for pointmap features. However, this approach has trade-offs, as shown in \Cref{tab:compare_pointmap_methods}: NoPoSplat has a significantly larger model (611M parameters vs. MVSplat's~\cite{chen2024mvsplat} 12M), and its rendering quality without test-time pose alignment (PSNR 24.70) is notably below MVSplat's (PSNR 26.39). While NoPoSplat achieves its best quality (PSNR 26.79) with test-time pose alignment, this step is slow, averaging ~2.84 seconds per instance, thereby limiting real-time applicability.

NoPoSplat's case illustrates that specialized heads for pointmap integration can lead to large models and require costly test-time optimizations for optimal performance. Motivated by these limitations, our \method instead distills pointmap priors via an efficient training loss, aiming for a better balance between rendering quality, model size, and inference speed.

\begin{table*}[t]
    \begin{center}
    \caption{\textbf{Quantitative comparison of NoPoSplat's variants and MVSplat.} Key metrics, model size, and inference time for NoPoSplat (with/without pose alignment) compared to the MVSplat baseline.}
    \vspace{0.2cm}
    \footnotesize
    \setlength{\tabcolsep}{2.5pt} 
    \begin{tabular}{@{}l ccc cc@{}}
    \toprule
    \multirow{2}{*}[-2pt]{\textbf{Method}} & \multicolumn{3}{c}{\textbf{Rendering Quality}} & \multirow{2}{*}{\textbf{Parameters (M)}$\downarrow$} & \multirow{2}{*}{\textbf{Time (ms)$\downarrow$}} \\
    \addlinespace[-12pt] \\
    \cmidrule{2-4}
    \addlinespace[-12pt] \\
    & PSNR$\uparrow$ & SSIM$\uparrow$ & LPIPS$\downarrow$ & & \\
    \midrule
    NoPoSplat (w/o pose align) & 24.70 & 0.818 & 0.145 & 611 & 2.8 \\
    NoPoSplat (w/ pose align) & 26.79 & 0.878 & 0.124 & 611 & 2840 \\
    MVSplat & 26.39 & 0.869 & 0.128 & 12 & 1.4 \\
    \bottomrule
    \end{tabular}
    \label{tab:compare_pointmap_methods}
    \end{center}
\end{table*}

\vspace{-0.1cm}
\section{Limitation and Societal Impacts}
\vspace{-0.1cm}
\label{sec:app_further_considerations}
\begin{figure}[t]
    \centering
    \includegraphics[width=\linewidth]{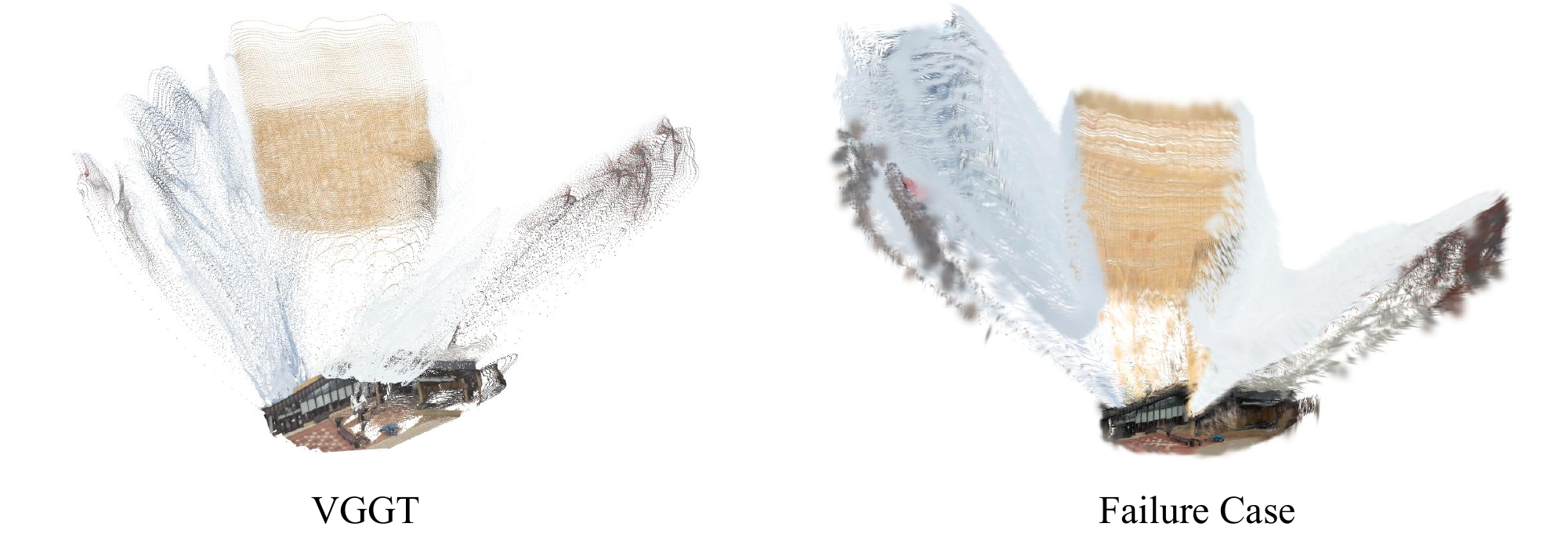}
    \vspace{-0.7cm}
    \caption{\textbf{Limitation: Pointmap Errors Propagate to 3DGS. }Left: VGGT pointmap with inaccuracies (e.g., for sky regions). Right: DepthSplat + \method's failure case 3D Gaussians showing propagated errors from the pointmap prior.}
    \label{fig:limitation}
\end{figure}
\textbf{Limitation: dependence on pointmap quality.}
As stated in the main paper, our \method's effectiveness is limited by the quality of the pre-trained pointmap model providing geometric priors. Errors in this pointmap can propagate via our \method, impacting the final 3DGS model's geometric quality. \Cref{fig:limitation} illustrates this dependency. The left side shows a VGGT pointmap with inaccuracies, while the right displays DepthSplat + \method's 3D Gaussians where these prior errors are visibly reflected. Thus, poor pointmap quality in challenging regions constrains our \method's performance. Leveraging improved pointmap models in the future could further enhance our approach.

\textbf{Broader societal impacts.}
Our work significantly improves 3D shape quality from feed-forward Gaussian Splatting models, presenting both opportunities and ethical considerations. Benefits include enhanced AR/VR experiences, accelerated 3D content creation, and improved cultural heritage preservation. However, the heightened realism also poses risks. A major risk is misuse for creating believable fake content; for example, high geometric quality could enable convincing fake 3D environments reconstructed from AI-generated images or videos, potentially fueling propaganda or scams and reducing digital trust. Furthermore, the synthetic nature of these models warrants caution in safety-critical applications, such as autonomous vehicle training, due to potential inaccuracies.

\begin{figure}[t]
    \centering
    \includegraphics[width=\linewidth]{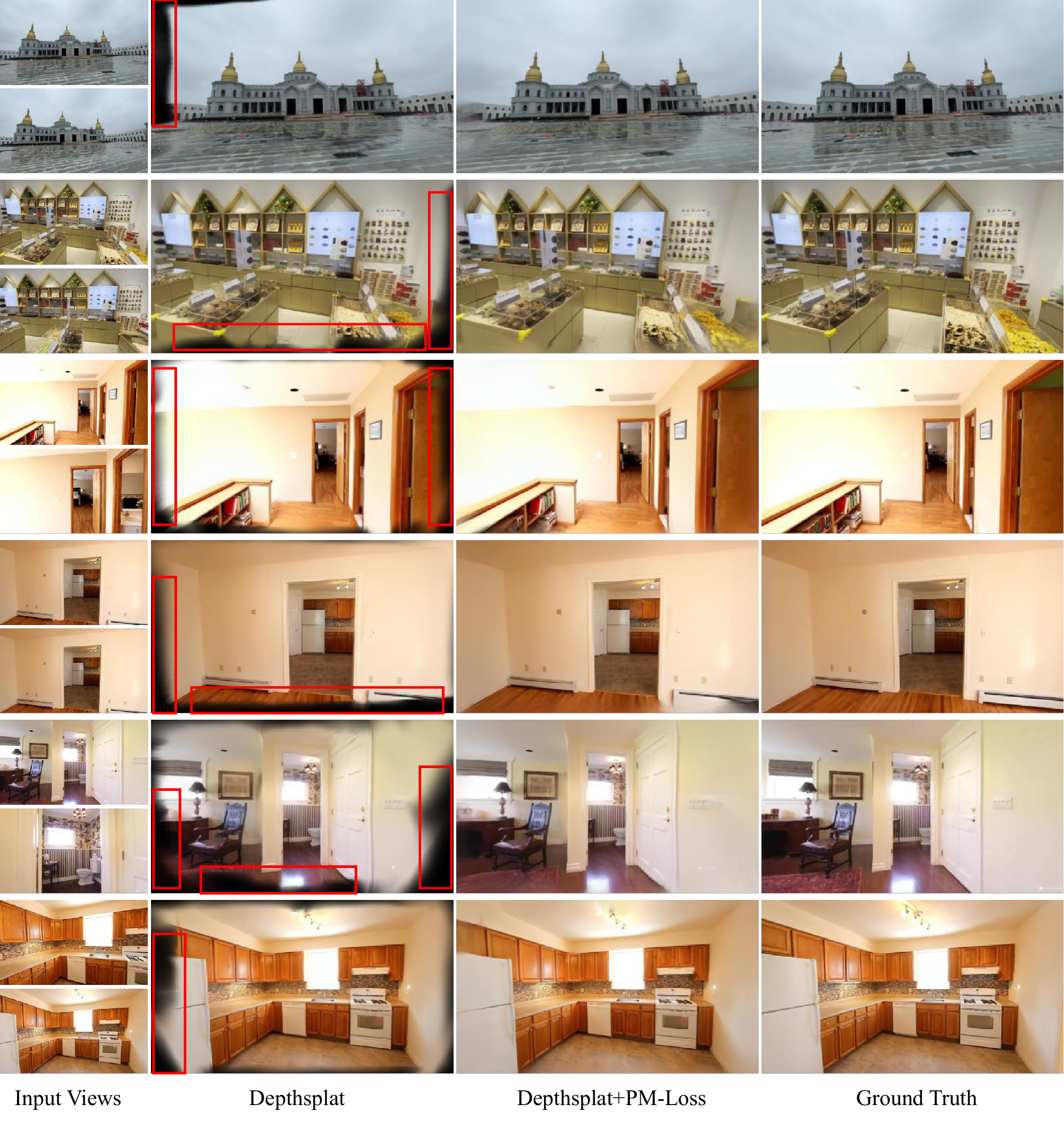}
    \vspace{-1.0cm}
    \caption{\textbf{More comparisons on DL3DV(top two rows) and RealEstate10K(bottom four rows) under the 2-view extrapolation setting. }Adding \method leads to significant improvements in rendering object boundaries.}
    \label{fig:more_render}
    \vspace{-0.5cm}
\end{figure}  

\begin{figure}[t]
    \centering
    \includegraphics[width=\linewidth]{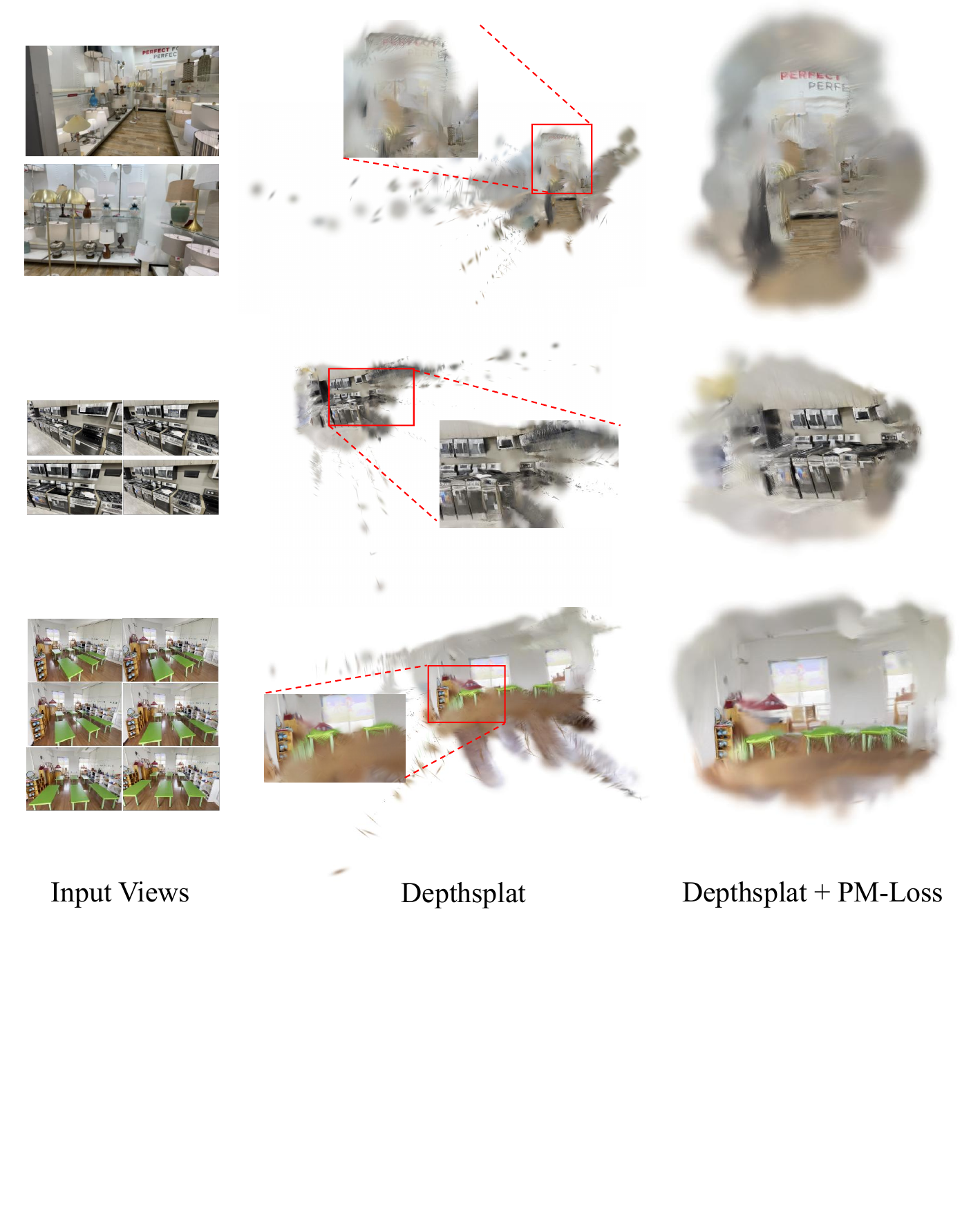}
    \vspace{-4.5cm}
    \caption{\textbf{More comparison of 3D Gaussians on DL3DV dataset. }Our method effectively constrains the 3D Gaussians, significantly reducing floating artifacts and noise near border.}
    \label{fig:more_gs_dl3dv}
    \vspace{-0.5cm}
\end{figure}

\vspace{-0.1cm}
\section{More Implementation Details}
\vspace{-0.1cm}
\label{sec:appendix_implementation_details}
\textbf{More baseline details.}
For our experiments involving DepthSplat~\cite{xu2024depthsplat}, it is important to note the specific version used. The results presented in this paper are based on the initial version of DepthSplat, which was publicly released in October 2024. We acknowledge that an updated version of the DepthSplat model architecture was made available in March 2025. However, all experiments reported herein were conducted using the aforementioned October 2024 release to maintain consistency with the experimental timeline of our work.

\textbf{More loss function details.}
Our total training loss, denoted as $L_{\text{total}}$, consists of two primary components: a rendering loss term, $L_{\text{render}}$, and our proposed geometric consistency loss, $L_{\text{PM}}$. The detailed formulation of $L_{\text{PM}}$ itself is provided in the main paper(Sec. 3.2). The overall loss is defined as:
\begin{equation}
    L_{\text{total}} = L_{\text{render}} + \lambda_{\text{PM}} L_{\text{PM}}
    \label{eq:appendix_total_loss}
\end{equation}
Here, $\lambda_{\text{PM}}$ is the weighting coefficient for $L_{\text{PM}}$. As specified in the main paper's Implementation Details section, we set $\lambda_{\text{PM}} = 0.005$.

The rendering loss, $L_{\text{render}}$, follows common practices established in prior works~\cite{chen2024mvsplat,xu2024depthsplat}. It combines a Mean Squared Error ($L_{\text{MSE}}$) term and an LPIPS ($L_{\text{LPIPS}}$) perceptual loss term:
\begin{equation}
    L_{\text{render}} = \lambda_{\text{MSE}} L_{\text{MSE}} + \lambda_{\text{LPIPS}} L_{\text{LPIPS}}
    \label{eq:appendix_render_loss}
\end{equation}
The $L_{\text{MSE}}$ term quantifies the pixel-wise squared differences between the rendered output and the ground-truth image. The $L_{\text{LPIPS}}$ term assesses perceptual similarity between the rendered and ground-truth images. We set the respective weights for these components as $\lambda_{\text{MSE}} = 1.0$ and $\lambda_{\text{LPIPS}} = 0.05$.

\textbf{More training details.}
We fine-tune all baseline models on the DL3DV dataset, starting from their publicly available pre-trained weights. To fairly evaluate our \method, we prepare two versions for each baseline: one fine-tuned with \method integrated, and a control version fine-tuned strictly following the baseline's original methodology. This isolates the impact of \method. While largely adhering to original training configurations, minor adjustments were made for single GPU fine-tuning.
For DepthSplat~\cite{xu2024depthsplat}, we use a batch size of 1, with the number of context views randomly chosen from 2-6 and target views fixed at 4 during training. For MVSplat~\cite{chen2024mvsplat}, the batch size is 12, with 2 fixed context views and 4 fixed target views. Code and pre-trained weights will be made available.

\textbf{More testing details.}
\label{app:more_testing_details}
Here we further details our Novel View Synthesis (NVS) extrapolation protocol, introduced in the main paper (Sec. 4.1 ). We use two context views and select target views outside their direct span. This setup rigorously assesses geometric quality, particularly at object edges, by ensuring genuine extrapolation that challenges models at scene peripheries.
Our specific sampling strategy, based on image sequence indices, adapts to dataset characteristics. For the RealEstate10K dataset, context views are separated by 75 image indices; we then sample 15 target views before the first context view and 15 after the second (30 total extrapolation views). For the DL3DV dataset, due to its larger inter-frame viewpoint changes, we use a context gap of 20 indices, sampling 10 target views before the first and 10 after the second context view (20 total extrapolation views).

\vspace{-0.3cm}
\section{More Visual Comparisons}
\vspace{-0.3cm}

This section presents supplementary visual comparisons to further demonstrate the impact of our \method. \cref{fig:more_render} showcases additional novel view synthesis results from the DL3DV and RealEstate10K datasets, particularly under the 2-view extrapolation setting, highlighting the improved rendering of object boundaries when our \method is applied. Furthermore, \cref{fig:more_gs_dl3dv} visualizes the quality of the reconstructed 3D Gaussians on the DL3DV dataset, illustrating the reduction in floating artifacts and border noise achieved by our approach.

%% file: main.bbl
\begin{thebibliography}{10}

\bibitem{kerbl20233d}
Bernhard Kerbl, Georgios Kopanas, Thomas Leimk{\"u}hler, and George Drettakis.
\newblock 3d gaussian splatting for real-time radiance field rendering.
\newblock {\em ACM Trans. Graph.}, 42(4):139--1, 2023.

\bibitem{huang20242d}
Binbin Huang, Zehao Yu, Anpei Chen, Andreas Geiger, and Shenghua Gao.
\newblock 2d gaussian splatting for geometrically accurate radiance fields.
\newblock In {\em ACM SIGGRAPH 2024 conference papers}, pages 1--11, 2024.

\bibitem{fan2024trim}
Lue Fan, Yuxue Yang, Minxing Li, Hongsheng Li, and Zhaoxiang Zhang.
\newblock Trim 3d gaussian splatting for accurate geometry representation.
\newblock {\em arXiv preprint arXiv:2406.07499}, 2024.

\bibitem{lyu20243dgsr}
Xiaoyang Lyu, Yang-Tian Sun, Yi-Hua Huang, Xiuzhe Wu, Ziyi Yang, Yilun Chen, Jiangmiao Pang, and Xiaojuan Qi.
\newblock 3dgsr: Implicit surface reconstruction with 3d gaussian splatting.
\newblock {\em ACM Transactions on Graphics (TOG)}, 43(6):1--12, 2024.

\bibitem{wolf2024surface}
Yaniv Wolf, Amit Bracha, and Ron Kimmel.
\newblock Surface reconstruction from gaussian splatting via novel stereo views.
\newblock {\em arXiv e-prints}, pages arXiv--2404, 2024.

\bibitem{yu2024gsdf}
Mulin Yu, Tao Lu, Linning Xu, Lihan Jiang, Yuanbo Xiangli, and Bo~Dai.
\newblock Gsdf: 3dgs meets sdf for improved rendering and reconstruction.
\newblock {\em arXiv preprint arXiv:2403.16964}, 2024.

\bibitem{charatan2024pixelsplat}
David Charatan, Sizhe~Lester Li, Andrea Tagliasacchi, and Vincent Sitzmann.
\newblock pixelsplat: 3d gaussian splats from image pairs for scalable generalizable 3d reconstruction.
\newblock In {\em CVPR}, pages 19457--19467, 2024.

\bibitem{chen2024mvsplat}
Yuedong Chen, Haofei Xu, Chuanxia Zheng, Bohan Zhuang, Marc Pollefeys, Andreas Geiger, Tat-Jen Cham, and Jianfei Cai.
\newblock Mvsplat: Efficient 3d gaussian splatting from sparse multi-view images.
\newblock In {\em ECCV}, pages 370--386. Springer, 2024.

\bibitem{xu2024depthsplat}
Haofei Xu, Songyou Peng, Fangjinhua Wang, Hermann Blum, Daniel Barath, Andreas Geiger, and Marc Pollefeys.
\newblock Depthsplat: Connecting gaussian splatting and depth.
\newblock In {\em CVPR}, 2025.

\bibitem{szymanowicz2024flash3d}
Stanislaw Szymanowicz, Eldar Insafutdinov, Chuanxia Zheng, Dylan Campbell, Joao~F Henriques, Christian Rupprecht, and Andrea Vedaldi.
\newblock Flash3d: Feed-forward generalisable 3d scene reconstruction from a single image.
\newblock {\em arXiv preprint arXiv:2406.04343}, 2024.

\bibitem{ziwen2024long}
Chen Ziwen, Hao Tan, Kai Zhang, Sai Bi, Fujun Luan, Yicong Hong, Li~Fuxin, and Zexiang Xu.
\newblock Long-lrm: Long-sequence large reconstruction model for wide-coverage gaussian splats.
\newblock {\em arXiv preprint arXiv:2410.12781}, 2024.

\bibitem{zhang2024gs}
Kai Zhang, Sai Bi, Hao Tan, Yuanbo Xiangli, Nanxuan Zhao, Kalyan Sunkavalli, and Zexiang Xu.
\newblock Gs-lrm: Large reconstruction model for 3d gaussian splatting.
\newblock In {\em ECCV}, pages 1--19. Springer, 2024.

\bibitem{wang2024freesplat}
Yunsong Wang, Tianxin Huang, Hanlin Chen, and Gim~Hee Lee.
\newblock Freesplat: Generalizable 3d gaussian splatting towards free view synthesis of indoor scenes.
\newblock {\em NeurIPS}, 37:107326--107349, 2024.

\bibitem{zhang2025transplat}
Chuanrui Zhang, Yingshuang Zou, Zhuoling Li, Minmin Yi, and Haoqian Wang.
\newblock Transplat: Generalizable 3d gaussian splatting from sparse multi-view images with transformers.
\newblock In {\em AAAI}, volume~39, pages 9869--9877, 2025.

\bibitem{ramamonjisoa2020predicting}
Michael Ramamonjisoa, Yuming Du, and Vincent Lepetit.
\newblock Predicting sharp and accurate occlusion boundaries in monocular depth estimation using displacement fields.
\newblock In {\em CVPR}, pages 14648--14657, 2020.

\bibitem{sun2023sc}
Libo Sun, Jia-Wang Bian, Huangying Zhan, Wei Yin, Ian Reid, and Chunhua Shen.
\newblock Sc-depthv3: Robust self-supervised monocular depth estimation for dynamic scenes.
\newblock {\em IEEE transactions on pattern analysis and machine intelligence}, 46(1):497--508, 2023.

\bibitem{wang2024dust3r}
Shuzhe Wang, Vincent Leroy, Yohann Cabon, Boris Chidlovskii, and Jerome Revaud.
\newblock Dust3r: Geometric 3d vision made easy.
\newblock In {\em CVPR}, pages 20697--20709, 2024.

\bibitem{yao2018mvsnet}
Yao Yao, Zixin Luo, Shiwei Li, Tian Fang, and Long Quan.
\newblock Mvsnet: Depth inference for unstructured multi-view stereo.
\newblock In {\em ECCV}, pages 767--783, 2018.

\bibitem{gu2020cascade}
Xiaodong Gu, Zhiwen Fan, Siyu Zhu, Zuozhuo Dai, Feitong Tan, and Ping Tan.
\newblock Cascade cost volume for high-resolution multi-view stereo and stereo matching.
\newblock In {\em CVPR}, pages 2495--2504, 2020.

\bibitem{yang2025fast3r}
Jianing Yang, Alexander Sax, Kevin~J Liang, Mikael Henaff, Hao Tang, Ang Cao, Joyce Chai, Franziska Meier, and Matt Feiszli.
\newblock Fast3r: Towards 3d reconstruction of 1000+ images in one forward pass.
\newblock In {\em CVPR}, 2025.

\bibitem{tang2024mv}
Zhenggang Tang, Yuchen Fan, Dilin Wang, Hongyu Xu, Rakesh Ranjan, Alexander Schwing, and Zhicheng Yan.
\newblock Mv-dust3r+: Single-stage scene reconstruction from sparse views in 2 seconds.
\newblock In {\em CVPR}, 2025.

\bibitem{zhang2024monst3r}
Junyi Zhang, Charles Herrmann, Junhwa Hur, Varun Jampani, Trevor Darrell, Forrester Cole, Deqing Sun, and Ming-Hsuan Yang.
\newblock Monst3r: A simple approach for estimating geometry in the presence of motion.
\newblock {\em arXiv preprint arXiv:2410.03825}, 2024.

\bibitem{wang2025vggt}
Jianyuan Wang, Minghao Chen, Nikita Karaev, Andrea Vedaldi, Christian Rupprecht, and David Novotny.
\newblock Vggt: Visual geometry grounded transformer.
\newblock In {\em CVPR}, 2025.

\bibitem{wang2024moge}
Ruicheng Wang, Sicheng Xu, Cassie Dai, Jianfeng Xiang, Yu~Deng, Xin Tong, and Jiaolong Yang.
\newblock Moge: Unlocking accurate monocular geometry estimation for open-domain images with optimal training supervision.
\newblock In {\em CVPR}, 2025.

\bibitem{wang20243d}
Hengyi Wang and Lourdes Agapito.
\newblock 3d reconstruction with spatial memory.
\newblock {\em arXiv preprint arXiv:2408.16061}, 2024.

\bibitem{smart2024splatt3r}
Brandon Smart, Chuanxia Zheng, Iro Laina, and Victor~Adrian Prisacariu.
\newblock Splatt3r: Zero-shot gaussian splatting from uncalibrated image pairs.
\newblock {\em arXiv preprint arXiv:2408.13912}, 2024.

\bibitem{ye2024no}
Botao Ye, Sifei Liu, Haofei Xu, Xueting Li, Marc Pollefeys, Ming-Hsuan Yang, and Songyou Peng.
\newblock No pose, no problem: Surprisingly simple 3d gaussian splats from sparse unposed images.
\newblock In {\em ICLR}, 2025.

\bibitem{yeshwanth2023scannet++}
Chandan Yeshwanth, Yueh-Cheng Liu, Matthias Nie{\ss}ner, and Angela Dai.
\newblock Scannet++: A high-fidelity dataset of 3d indoor scenes.
\newblock In {\em ICCV}, pages 12--22, 2023.

\bibitem{zhou2018stereo}
Tinghui Zhou, Richard Tucker, John Flynn, Graham Fyffe, and Noah Snavely.
\newblock Stereo magnification: learning view synthesis using multiplane images.
\newblock {\em ACM Transactions on Graphics (TOG)}, 37(4):1--12, 2018.

\bibitem{ling2024dl3dv}
Lu~Ling, Yichen Sheng, Zhi Tu, Wentian Zhao, Cheng Xin, Kun Wan, Lantao Yu, Qianyu Guo, Zixun Yu, Yawen Lu, et~al.
\newblock Dl3dv-10k: A large-scale scene dataset for deep learning-based 3d vision.
\newblock In {\em CVPR}, pages 22160--22169, 2024.

\bibitem{chen2023view}
Shenchang~Eric Chen and Lance Williams.
\newblock View interpolation for image synthesis.
\newblock In {\em Proceedings of the 20th Annual Conference on Computer Graphics and Interactive Techniques}, SIGGRAPH '93, page 279–288, New York, NY, USA, 1993. Association for Computing Machinery.

\bibitem{seitz1996view}
Steven~M Seitz and Charles~R Dyer.
\newblock View morphing.
\newblock In {\em Proceedings of the 23rd annual conference on Computer graphics and interactive techniques}, pages 21--30, 1996.

\bibitem{mildenhall2020nerfrepresentingscenesneural}
Ben Mildenhall, Pratul~P Srinivasan, Matthew Tancik, Jonathan~T Barron, Ravi Ramamoorthi, and Ren Ng.
\newblock Nerf: Representing scenes as neural radiance fields for view synthesis.
\newblock {\em Communications of the ACM}, 65(1):99--106, 2021.

\bibitem{chung2024depthregularizedoptimization3dgaussian}
Jaeyoung Chung, Jeongtaek Oh, and Kyoung~Mu Lee.
\newblock Depth-regularized optimization for 3d gaussian splatting in few-shot images, 2024.

\bibitem{zhu2023FSGS}
Zehao Zhu, Zhiwen Fan, Yifan Jiang, and Zhangyang Wang.
\newblock Fsgs: Real-time few-shot view synthesis using gaussian splatting.
\newblock In {\em ECCV}, 2024.

\bibitem{depth_anything_v1}
Lihe Yang, Bingyi Kang, Zilong Huang, Xiaogang Xu, Jiashi Feng, and Hengshuang Zhao.
\newblock Depth anything: Unleashing the power of large-scale unlabeled data.
\newblock In {\em CVPR}, 2024.

\bibitem{depth_anything_v2}
Lihe Yang, Bingyi Kang, Zilong Huang, Zhen Zhao, Xiaogang Xu, Jiashi Feng, and Hengshuang Zhao.
\newblock Depth anything v2.
\newblock In {\em NeurIPS}, 2024.

\bibitem{wang2025freesplat++}
Yunsong Wang, Tianxin Huang, Hanlin Chen, and Gim~Hee Lee.
\newblock Freesplat++: Generalizable 3d gaussian splatting for efficient indoor scene reconstruction, 2025.

\bibitem{min2024epipolarfree}
Zhiyuan Min, Yawei Luo, Jianwen Sun, and Yi~Yang.
\newblock Epipolar-free 3d gaussian splatting for generalizable novel view synthesis, 2024.

\bibitem{fei2024pixelgaussian}
Xin Fei, Wenzhao Zheng, Yueqi Duan, Wei Zhan, Masayoshi Tomizuka, Kurt Keutzer, and Jiwen Lu.
\newblock Pixelgaussian: Generalizable 3d gaussian reconstruction from arbitrary views, 2024.

\bibitem{kang2025selfsplat}
Gyeongjin Kang, Jisang Yoo, Jihyeon Park, Seungtae Nam, Hyeonsoo Im, Sangheon Shin, Sangpil Kim, and Eunbyung Park.
\newblock Selfsplat: Pose-free and 3d prior-free generalizable 3d gaussian splatting.
\newblock In {\em CVPR}, 2025.

\bibitem{wang2025zpressor}
Weijie Wang, Donny~Y. Chen, Zeyu Zhang, Duochao Shi, Akide Liu, and Bohan Zhuang.
\newblock Zpressor: Bottleneck-aware compression for scalable feed-forward 3dgs, 2025.

\bibitem{chen2024mvsplat360}
Yuedong Chen, Chuanxia Zheng, Haofei Xu, Bohan Zhuang, Andrea Vedaldi, Tat-Jen Cham, and Jianfei Cai.
\newblock Mvsplat360: Feed-forward 360 scene synthesis from sparse views.
\newblock {\em NeurIPS}, 37:107064--107086, 2024.

\bibitem{mast3r_arxiv24}
Vincent Leroy, Yohann Cabon, and Jerome Revaud.
\newblock Grounding image matching in 3d with mast3r, 2024.

\bibitem{zhang2025flare}
Shangzhan Zhang, Jianyuan Wang, Yinghao Xu, Nan Xue, Christian Rupprecht, Xiaowei Zhou, Yujun Shen, and Gordon Wetzstein.
\newblock Flare: Feed-forward geometry, appearance and camera estimation from uncalibrated sparse views.
\newblock In {\em CVPR}, 2025.

\bibitem{cut3r}
Qianqian Wang, Yifei Zhang, Aleksander Holynski, Alexei~A. Efros, and Angjoo Kanazawa.
\newblock Continuous 3d perception model with persistent state.
\newblock In {\em CVPR}, 2025.

\bibitem{vit}
Alexey Dosovitskiy, Lucas Beyer, Alexander Kolesnikov, Dirk Weissenborn, Xiaohua Zhai, Thomas Unterthiner, Mostafa Dehghani, Matthias Minderer, Georg Heigold, Sylvain Gelly, Jakob Uszkoreit, and Neil Houlsby.
\newblock An image is worth 16x16 words: Transformers for image recognition at scale.
\newblock In {\em ICLR}, 2021.

\bibitem{besl1992method}
Paul~J Besl and Neil~D McKay.
\newblock Method for registration of 3-d shapes.
\newblock In {\em Sensor fusion IV: control paradigms and data structures}, volume 1611, pages 586--606. Spie, 1992.

\bibitem{88573}
S.~Umeyama.
\newblock Least-squares estimation of transformation parameters between two point patterns.
\newblock {\em IEEE Transactions on Pattern Analysis and Machine Intelligence}, 13(4):376--380, 1991.

\bibitem{jensen2014large}
Rasmus Jensen, Anders Dahl, George Vogiatzis, Engin Tola, and Henrik Aan{\ae}s.
\newblock Large scale multi-view stereopsis evaluation.
\newblock In {\em CVPR}, pages 406--413, 2014.

\bibitem{chen2023matchnerf}
Yuedong Chen, Haofei Xu, Qianyi Wu, Chuanxia Zheng, Tat-Jen Cham, and Jianfei Cai.
\newblock Explicit correspondence matching for generalizable neural radiance fields, 2023.

\bibitem{ravi2020pytorch3d}
Nikhila Ravi, Jeremy Reizenstein, David Novotny, Taylor Gordon, Wan-Yen Lo, Justin Johnson, and Georgia Gkioxari.
\newblock Accelerating 3d deep learning with pytorch3d, 2020.

\end{thebibliography}
